\documentclass{article}

\PassOptionsToPackage{numbers, compress}{natbib}

     \usepackage[preprint]{neurips_2019}

\usepackage[utf8]{inputenc} 
\usepackage[T1]{fontenc}    
\usepackage{hyperref}       
\usepackage{url}            
\usepackage{booktabs}       
\usepackage{amsfonts}       
\usepackage{nicefrac}       
\usepackage{microtype}      
\usepackage{graphicx}
\usepackage{subfig}
\usepackage{comment}
\title{Faster and Accurate Classification for JPEG2000 
	Compressed Images in Networked Applications}

\author{%
	Lahiru D. Chamain, Zhi Ding
	\\
	Department of Electrical \& Computer Engineering \\
	University of California\\
	Davis, CA 95616 \\
	\texttt{\{hdchamain, zding\}@ucdavis.edu}
}
\begin{document}
	\maketitle
	\vspace*{-5mm}
	\begin{abstract}
		JPEG2000 (j2k) is a highly popular format for image and video compression. 
		With the rapidly growing applications of cloud based image 
		classification, most existing j2k-compatible schemes
		would stream compressed color images from the source 
		before reconstruction at the processing center as inputs to deep CNNs.
		We propose to remove the computationally costly reconstruction 
		step by training a deep CNN image classifier using
		the CDF 9/7 Discrete Wavelet Transformed (DWT) coefficients 
		directly extracted from j2k-compressed images.
		We demonstrate additional computation savings by utilizing
		shallower CNN to achieve classification of good accuracy 
		in the DWT domain. Furthermore, we show that
		traditional augmentation
		transforms such as flipping/shifting are ineffective
		in the DWT domain and present 
		different augmentation transformations to achieve
		more accurate classification without any additional cost.
		This way, faster and more accurate classification is possible 
		for j2k encoded images without image reconstruction.
		Through experiments on CIFAR-10 and Tiny ImageNet  data sets, 
		we show that the performance of the proposed solution is consistent 
		for image transmission over limited channel bandwidth.
	\end{abstract}\vspace*{-5mm}
	\section{Introduction}
	Artificial intelligence (AI), IoT and 5G are among the
	most exciting and game-changing technologies today, 
	fueling 
	new generations of technical solutions to some of the world’s biggest problems. Image/video recognition 
	in a networked environment is among
	the top AI applications for years to come. 
	Given its rapidly growing
	popularity and usage for visual applications, 
	JPEG2000 is playing an increasingly vital role
	in cyber intensive and autonomous  systems.
	In this work, we explore new and better ways to 
	exploit and optimize JPEG2000 (j2k) encoding
	in critical AI tasks such as image and 
	video recognition.
	
\begin{figure}[!htbp]
		\vspace*{-3mm}
		\centering
		\includegraphics[width=130mm,keepaspectratio,]{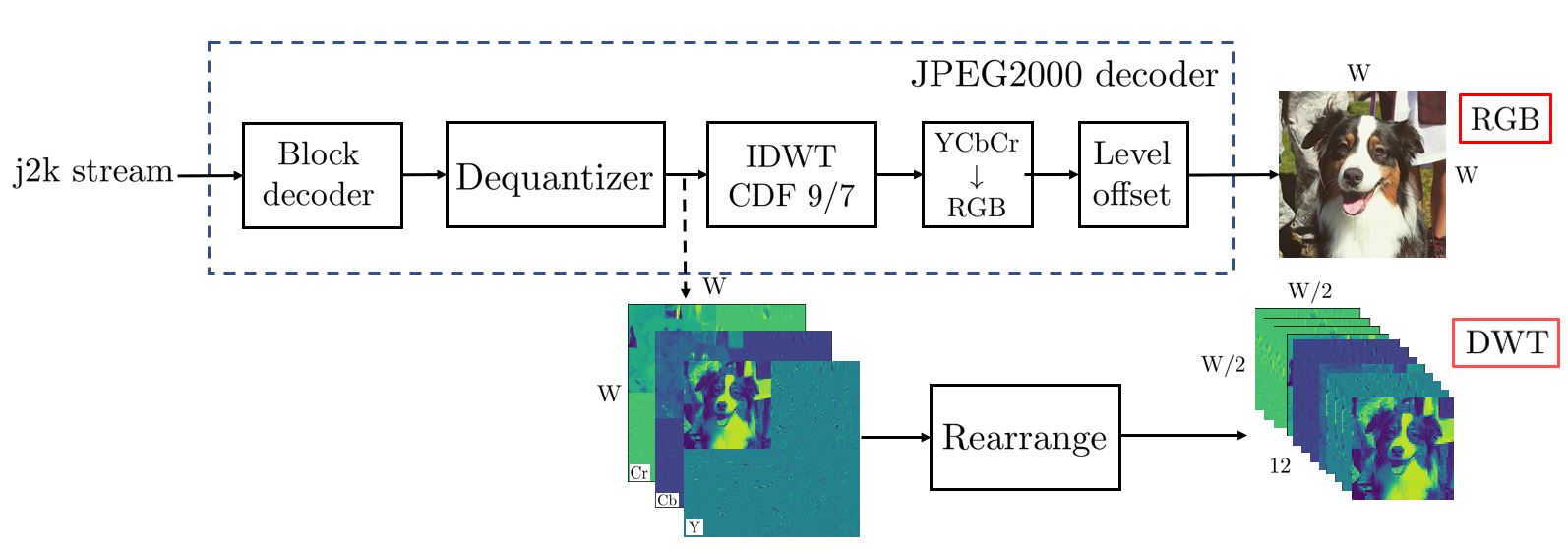}
		\caption{We adopt a standard JPEG2000 decoder. 
			After dequantization, we harvest 
			the CDF 9/7 coefficients by breaking into the JPEG2000 
			decoder. For illustration, we show a level-1 DWT 
			compressed image. We rearrange the CDF 9/7
			coefficients as inputs to the deep CNN. }
		\label{figj2k}
		\vspace*{-3mm}
	\end{figure}

	In a wide variety of application scenarios involving
	low complexity IoT and networked sensors, 
	traditional AI functionalities 
	often rely on cloud computing to handle 
	high complexity processing
	tasks such as image recognition. 
	There is clearly a trade-off among computation
	complexity, network payload, and performance in
	terms of accuracy. 
	In cloud based AI image applications, neural networks are
	developed and trained in the cloud or on a server
	to which user images are sent over a channel.
	To obtain classification labels, servers input the received
	images to its trained neural networks. 
	In order to conserve limited channel bandwidth and storage capacity, 
	source devices often encode and compress the images 
	before transmitting to the cloud by utilizing standardized
	compression techniques such as JPEG2000.
	Because most neural networks are designed
	to classify images in the spatial RBG domain, 
	the cloud currently receives and decodes the 
	compressed j2k images back into the RGB domain before
	forwarding them to trained neural networks for further processing,
	as illustrated in the top part of the Figure~\ref{figj2k}. 
	Thus, a natural question arises is to how to achieve
	faster training and inference with improved accuracy 
	in a cloud based image classification under bandwidth, storage and computation constraints. 
	
	In this work, we study the trade-off among computation complexity,
	accuracy and compression in a cyber-based image recognition
	system consisting of low complexity cameras, cloud servers,
	and band-limited communication links that connect them.

	We claim that the conventional use of image reconstruction is 
	unnecessary for JPEG2000 encoded classification by constructing
	and training a deep CNN model with the DWT coefficients
	transmitted in standard j2k stream. 
	See the bottom part of the Figure~\ref{figj2k}. 
	This result is consistent with the work of 
	\cite{gueguen2018faster} for JPEG encoded images. Furthermore,
	we establish that more accurate classification is also possible by deploying shallower models to benefit from
	faster training and classification 
	in comparison to models trained fo spatial RGB image 
	inputs. 
	
\begin{figure}[!htbp]
		\centering
		\vspace*{-5mm}
		\hspace*{-4mm}
		\subfloat[inference ]{{\includegraphics[width=8cm]{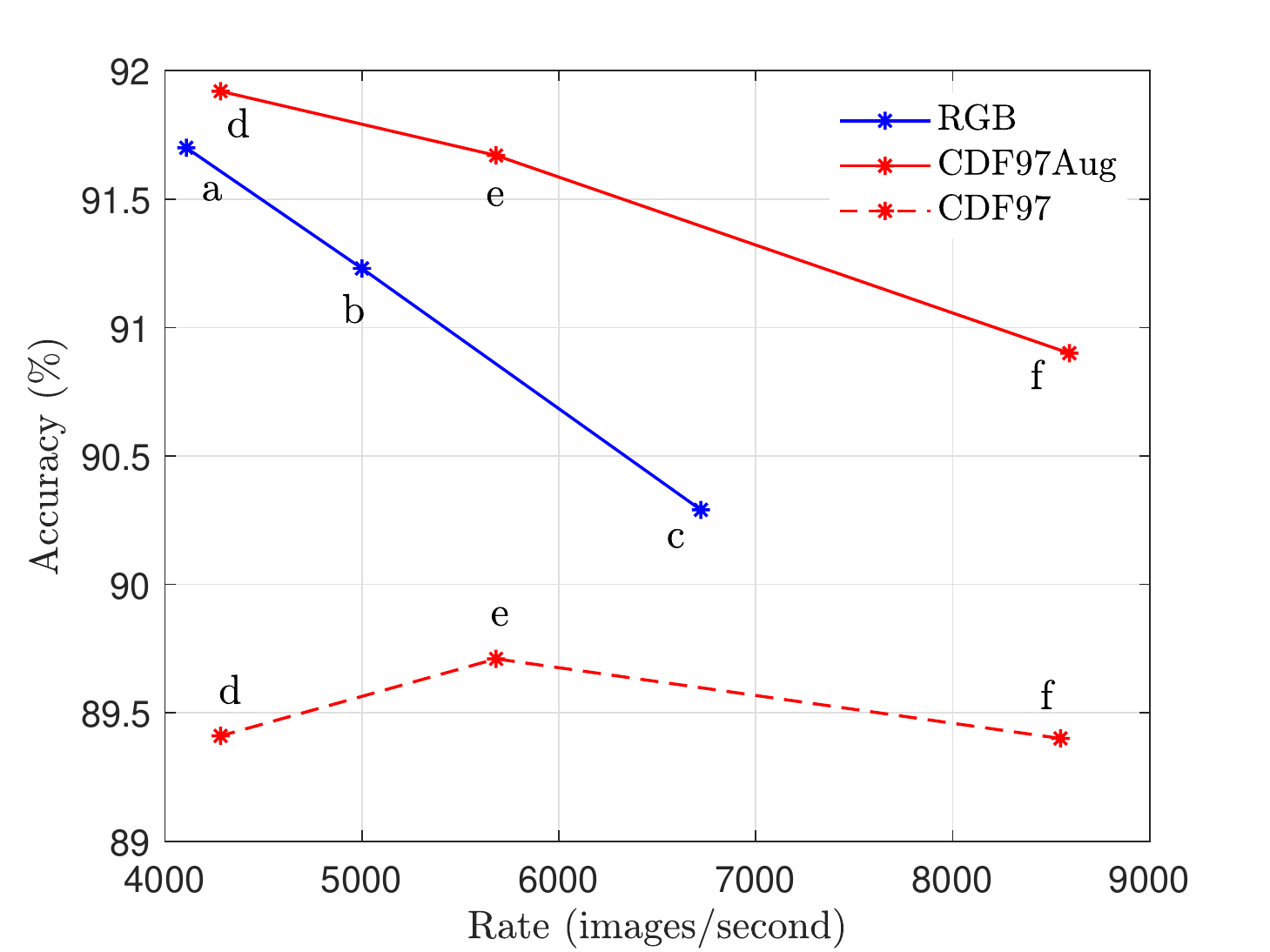}}}\hspace*{-6mm}
		\subfloat[training]{{\includegraphics[width=8cm]{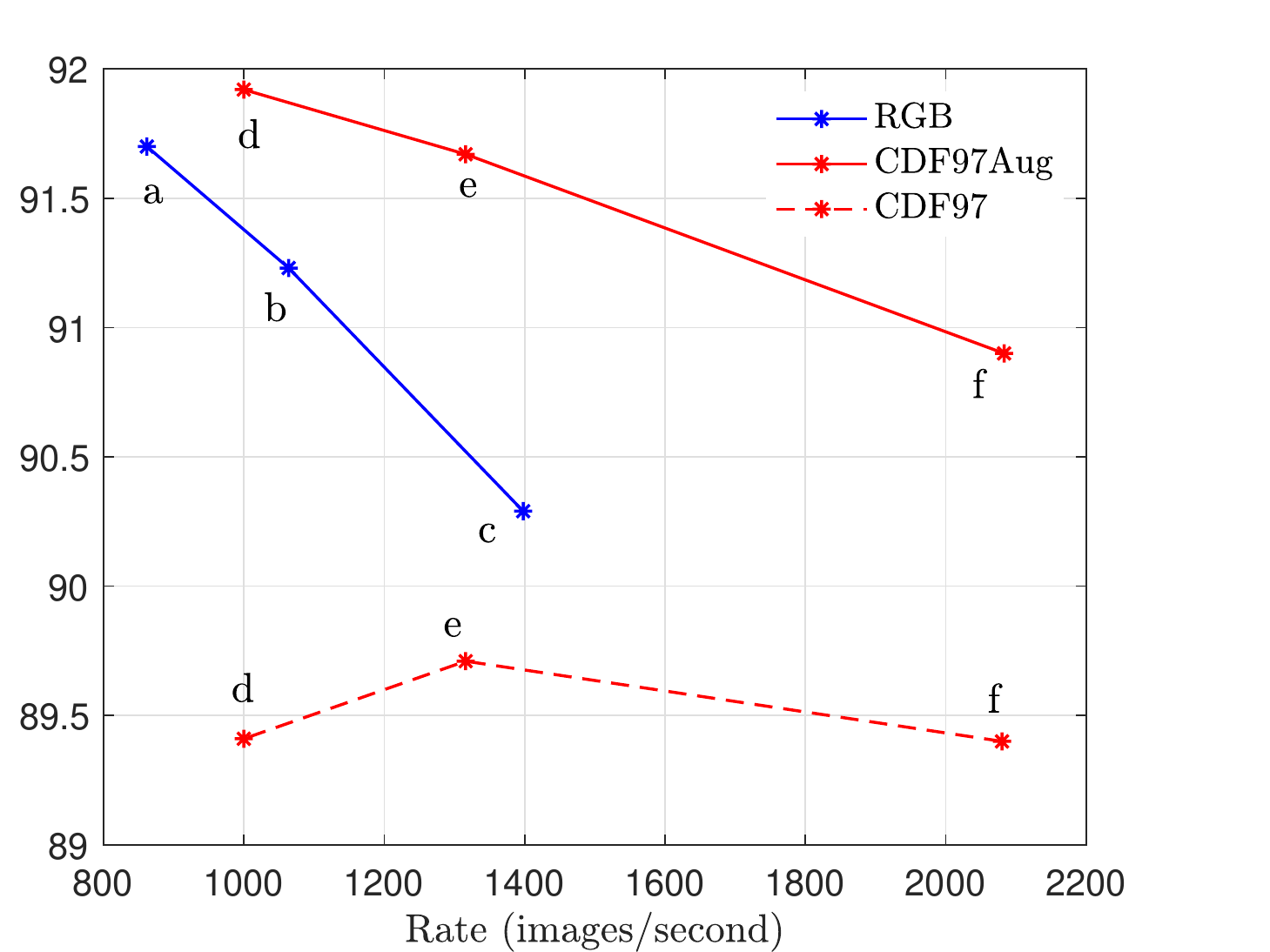}}}
		\caption{(a). Test accuracy vs inference speed 
			for the CIFAR-10 data set. The blue lines represent results using reconstructed 
			RGB images. Red curves are results using
			DWT coefficients extracted from JPEG2000 codec: 
			dashed lines ``CDF97'' from regular augmentation
			and solid lines ``CDF97Aug'' from our 
			proposed augmentation. (b). Test error vs training speed/epoch. 
			Here rate is the number of images that go through the model 
			in each epoch.  The proposed model delivers fast
			and accurate classification for both training and inference.}
		\label{figtesttrain}
		\vspace*{-5mm}
	\end{figure}

	Figure~\ref{figtesttrain} describes our first set of
	results. To achieve these results,
	we introduce novel augmentation transforms 
	in the DWT domain instead of replicating conventional 
	spatial image transformations that lead to accuracy degradation.
	
	We explore the consistency of the proposed solution over band limited channels by changing the compression ratio of the JPEG2000 codec and observe that the accuracy improvement of DWT domain is increasing compared to RGB domain. Finally we show that for the training of deep CNN models for band limited channels can use the pre-trained models for no compression case. This way we can cut down the training time of the models by 75\% while leading to improved accuracy. This observation is consistent with using pre-trained models in RGB domain for band limited scenarios.
	
	\section{Related Works}
	Image classification in spectral domains has been studied by
	the machine learning (ML) community over the past few years. 
	The compact representation of the images in spectral domains
	promises faster classification \cite{gueguen2018faster,zou2014high,williams2016,fu2016using}. 
	On the other hand, image compression for faster 
	transmission or smaller storage has long been applied to spectral transformation to separate and remove 
	information redundancy. These compact images 
	require less bandwidth and storage in cloud based AI systems. 
	A recent work \cite{chamain2018QuanNet} discusses how to 
	learning can be applied to more efficiently
	quantize DB1 wavelet coefficients to reduce image size 
	without compromising classification accuracy. Another recent work \cite{jiang2018end} suggests an encoder decoder framework to learn compact representation before JPEG2000 compression. 
	
	The authors of \cite{gueguen2018faster} applied
	existing JPEG codec to directly extract the Discrete Cosine transformed (DCT) coefficients for classifying imageNet~\cite{ILSVRC15} dataset. They claim faster
	classification by reducing some blocks of the ResNet stack, 
	together with the time saved from not reconstructing RGB 
	images before ResNet. Similar to this approach on JPEG, this paper uses the standard JPEG2000 codec to extract 
	DWT coefficients for classification. We show not only faster
	but also more accurate classification results 
	even before considering the reconstruction savings.
	Our results are from experiments on 
	both CIFAR-10~\cite{krizhevsky2012} 
	as well as Tiny ImageNet (a subset of ImageNet~\cite{ILSVRC15})  datasets. 
	
	To the best of authors' knowledge, this is the first 
	reported work that successfully applies 
	DWT coefficients extracted from within the JPEG2000 decoder 
	for image classification. 
	The authors of \cite{torfason2018towards} demonstrated a 
	similar concept of skipping reconstruction in ResNet classifier
	for a less common convolution encoder/decoder 
	instead of the widely popular JPEG2000 encoder/decoder in
	multimedia applications. 
	
	JPEG2000 uses CDF 9/7 wavelet as the mother wavelet 
	in the discrete wavelet transformation (DWT). 
	Despite the considerable volume of works on
	the use of DWT coefficients for image compression
	(e.g., \cite{williams2016,chamain2018QuanNet,fujieda2017,levinskis2013,kang2017}), there are only a handful of published works
	that use CDF 9/7 wavelets for classification. For example, 
	the authors of 
	\cite{williams2016,chamain2018QuanNet,levinskis2013} uses DB1, mostly known as `Harr' wavelet for DWT calculation.
	

Recent works \cite{gueguen2018faster},\cite{chamain2018QuanNet} 
and \cite{torfason2018towards} have respectively 
shown that reconstruction 
of RGB images from DCT, DB1 and a custom compressed coefficients
is unnecessary for classification purposes. 
The authors of \cite{chamain2018QuanNet} reported a
classification accuracy of 88.9\% for CIFAR-10 on a modified ResNet-20 by using unquantized CDF 9/7 DWT coefficients as 
CNN inputs. Furthermore they concluded that DB1 coefficients 
can achieve better accuracy over CDF 9/7 wavelets using the 
same network. 
However,  none of the previous works addressed the extension of regular  augmentation techniques in transform domains such as 
DCT or DWT. 
In this work, we achieved over 91.9\% accuracy for CDF 9/7 coefficients extracted from JPEG2000 codec for the same dataset
by applying our proposed augmentation techniques.
	
In another recent work \cite{chamain2018QuanNet}, 
the effect of the limited channel bandwidth in cloud based 
classification has been investigated for images in
DB1 and RGB domains. In this paper, our work
based on the CDF 9/7 coefficients of JPEG2000 
obtains consistent results and similar conclusions.

\section{Classification of JPEG2000 Compressed Images}

	\subsection{JPEG2000 Compression Code}
	JPEG2000 codec offers two main compression paths, 
reversible and irreversible compression 
\cite{taubman2012jpeg2000}. The reversible path uses 
CDF 5/3 wavelets for DWT calculation whereas
the irreversible path is lossy and uses CDF 9/7 wavelets. 
This paper focuses on the lossy irreversible 
compression. 
	
In the JPEG2000 encoder, an RGB image with integer pixel
values from 0 to 255 is first given a level offset of 
128 to shift the intensity distribution center to 0. 
The offset image is then converted to YCbCr color space 
by applying a linear mapping to facilitate
compression. From the YCbCr color space, 
CDF 9/7 wavelet coefficients are generated
from a DWT transformation. 
The resulting DWT coefficients are then quantized with 
a deadzone quantizer as explained in \cite{chamain2018QuanNet} before sending to the block encoder. 
In the block encoder, the DWT coefficients are
encoded using arithmetic
code and efficiently arranged with EBCOT algorithm \cite{taubman2000high} to produce the eventual j2k stream.
	
The decoder, as shown in Figure~\ref{figj2k},
extracts the CDF 9/7 coefficients using the 
dequantizing block which multiplies the received DWT
oefficients with the same step size used by the encoder during
quantization. In our experiments, we modified the open source `C' codes of the OpenJPEG project \footnote{http://www.openjpeg.org/}
to extract DWT coefficients and to generate RGB images.

\subsection{ResNet Inputs}
The bottom part of the Figure~\ref{figj2k} shows the process of
adapting DWT coefficients extracted directly from the decoder 
to ResNet \cite{he2016deep}. Similar to \cite{chamain2018QuanNet} three level-1 DWT 
channels corresponding to Y, Cb, and Cr are stacked in to a tensor of 12 sub bands each with the dimension of half of the RGB image.

\subsection{Image Augmentation}\label{secAug}
\begin{figure}[!htbp]
		\centering
		\vspace*{-3mm}
		\subfloat[original ]{{\includegraphics[width=3.2cm]{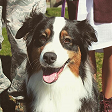} }}
		\subfloat[flipped in RGB]{{\includegraphics[width=3.2cm]{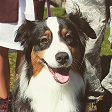} }}
		\subfloat[fillped in DWT]{{\includegraphics[width=3.2cm]{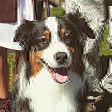} }}
		\subfloat[proposed]{{\includegraphics[width=3.2cm]{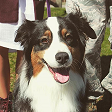} }}
		\caption{(a) Original image. `Brownie' of size $112\times112$. (b) shows the horizontally flipped image in RGB domain. But at the training, mini batches are in DWT domain. (c) shows the result of using the same transformation in DWT domain and reconstructed in RGB for visualization. (d) horizontally flipped in DWT domain with proposed augmentation transform and reconstructed in RGB for visualization. (b) and (d) are exactly same proving the effectiveness of the proposed method. (c) is distorted implying the ineffectiveness of the conventional augmentation transforms.}
		\label{figaug}
		\vspace*{-3mm}
	\end{figure}

Image augmentation is an essential step in deep CNN models 
to combat over fitting. By training
with augmented images one can generalize the CNN model to 
classify unseen images during training. 
In each mini batch, augmentation can provide random 
transformations such as horizontal flipping, rotation, 
vertical and horizontal 
shifting. These transformations are meaningful in the 
spatial domain like RGB or YCbCr. However,
inputs in DWT or other transform domains, these conventional transformations do not have physical meaning and have
proven ineffective. Figure \ref{figaug} shows a clear
distortion in (c) as a result of incorrectly flipped 
high frequency sub-bands.  

We address this problem by proposing
the following augmentation transforms. 
Let $n$ be the image dimension. Define
$X_{s}\in\mathcal{R}^{n\times n}$ and $X_{w_A}\in\mathcal{R}^{n\times n}$ where $X_{w_A}$ is 
the DWT of $X_{s}$ and $A$ is the DWT matrix of a wavelet 
$a$. 
Now we can write $X_{w_A}$ as separable column and row transformations (See \cite{gonzalez2002digital} for more details).
	\begin{equation}\label{DWT}
	X_{w_A} = A^{T}X_{s}A.
	\end{equation}
We define the regular spatial domain transformation $H$ as
	\begin{equation}
	X_{s_H} = X_{s}H
	\end{equation}
	where $X_{s_H}$ is the augmented  $X_{s}$ in the spatial domain. We suggest an alternative transformation $\acute{H}$ defined as
	\begin{equation}
	\acute{H} = A^{-1}HA
	\end{equation}
which replaces $H$ in augmentation in DWT domain. (See Section~\ref{secEffA} for the formulation of A).

\subsection{Computation Speed and Accuracy}
We achieve faster training and inference in
three ways. 
First, we save computation cost
by not reconstructing (Reconstruction gain).
Second, shallower models are sufficient for image classification in spectral domain and 
lead to less computations and faster training. 
Third, by applying specialized DWT 
augmentation, we improve classification 
accuracy. 
The first two methods are consistent
with known results for JPEG
as faster classification can be achieved with fewer ResNet blocks \cite{gueguen2018faster}.
	
\subsection{Reconstruction Gain}\label{recongain}

Deep CNNs such as ResNet thrive on RGB inputs and have been shown effective with 
augmentation transforms like flipping, random cropping and rotations etc. In cloud based image classification,
the receiver node receives
j2k stream of each inference image. 
We demonstrate that the JPEG decoder
steps after the dequantizer as shown in Figure \ref{figj2k} are unnecessary to achieve the same level accuracy as reconstructed RGB images. 
Hence the computation required for IDWT,
YCbCR to RGB conversion, and the level 
offset can be omitted. We call this 
computation/time saving 
``reconstruction gain' which is dominated by the IDWT of CDF 9/7 wavelets in JPEG2000. 
The reconstruction gain amounts to over 80\% 
decoding time of the OpenJPEG CPU implementation. 
See Table \ref{recongainTable}.
	
\subsection{Shallow Models}
Deep CNN models with fewer number 
of residual blocks ($B(nf_s)$) are sufficient  to achieve a given classification accuracy
in DWT domain with CDF 9/7 wavelets in
comparison to classification in the
RGB domain. 
This can be seen by experimenting with 6 ResNet models parameterized as shown in Table \ref{resnettable}. On the other hand, the width of the RGB inputs
is 2 times larger than the CDF 9/7 inputs 
(for level 1 DWT),
but DWT requires more convolution filters 
for each layer to compensate the larger 
input depth (12 channels) in CDF 9/7 compared to the 3 RGB  channels.
	
\subsection{Efficient Augmentation}\label{secEffA}

The basis for the proposed augmentation
transforms on DWT domain in Section
\ref{secAug} is the ability to represent 
DWT operation 
as an invertible linear operation as shown
in Eq.~(\ref{DWT}). We formulated the matrix $A\in\mathcal{R}^{n\times n}$ for CDF 9/7 wavelets as follows.
	
The lifting implementation of the 1-Dimensional
DWT of CDF 9/7 wavelet consists of two predictions($ P_1 $, $P_2$), each followed by an update function ($U_1$ and $U_2$). 
Then the resulting matrix is de-interleaved ($S$) to form high and low frequency components. In OpenJPEG, this implementations consists of five ``for'' 
loops for each operation.
	
We observed that this DWT matrix operation can
be implemented as a matrix multiplication of
three sub-functions: predictions, updates and de-interleaving. It can be denoted as 
	\begin{equation}
	A = P_1 U_1 P_2 U_2 S.
	\end{equation}
Matrix $A$ and its inverse calculated this way\footnote{python code for this transformations is available at\\ \hyperlink{cdf9/7 matrix implementation}{https://drive.google.com/open?id=16ZLKu107TSrnWuFi7Fw80MTT3Ck4008l}}
can be stored at the start of CNN training.
	
With this implementation we could reduce the DWT conversion time of 10,000 images of size $32\times32\times3$ from 4 minutes to a mere
0.5 of a second on an INTEL $ 7^{th} $ GEN CORE i7-7700HQ CPU. 

\subsection{Bandwidth Constrained Cloud Image Classification}
In cloud based image classification, 
the available bandwidth can critically affect 
the classification accuracy. The authors of \cite{chamain2018QuanNet} discussed 
this effect in more details. 
In JPEG2000 encoding, the precision of DWT
coefficients correspond to quality as
higher quality layers offer more precision
during encoding. When a compression 
parameter `$r$' is used by the encoder, 
it changes the number of higher quality level 
bits proportionally. 
This is equivalent to using larger step sizes in the dead zone quantizer.
	
We can test how robustly
deep CNN image classifiers would respond
to JPEG2000 
encoded images over channels with different bandwidths. 
We can easily observe such results by 
training a deep CNN model on the decoded DWT coefficients of images encoded at different
compression ratio $r$. To confirm that
image reconstruction in 
deep CNN image classification of j2k 
is unnecessary, the testing accuracy of the DWT domain inputs should respond similarly
to RGB image classifiers 
against a given bandwidth.
	
\section{Experiments and Results}

\subsection{Pre-Training and Fine-Tuning}
	
In training a cloud based ML application for different available bandwidths,
one intuitive way
is to train a base model for maximum allowable bandwidth or using unquantized inputs.
We then fine-tune the model by using the base model as the pre-trained model. 
This approach can reduce time needed in training different models for different
levels of available bandwidths,
as pre-training and fine-tuning can 
ease the convergence of the model 
to a higher accuracy level. 

	
\subsection{Results}
In the initial experiments,
we used CIFAR-10 dataset which consists of 50,000 training images and 10000 testing images of size 32$\times$32 belonging to 10 classes. We then repeated the experiments on Tiny ImageNet, as a subset of ImageNet dataset 
\cite{ILSVRC15}. Tiny ImageNet consists of RGB images of size 64 $\times$ 64 belonging to 200 classes, each class with
1300 training images and 50 validation images.
For both CIFAR-10 and Tiny ImageNet, we used `Adam' as the optimizer.
	
To compare the classification accuracy in
RGB and DWT domains, we start by
encoding original (source) images in the training set using JPEG2000 encoder at compression ratio $r$ to generate 
j2k streams for each image. We then 
decode the j2k 
streams into RGB images for RGB domain inputs 
and harvest their DWT values 
inside the codec for DWT domain) inputs as 
explained
 in Fig.~\ref{figj2k}.
	
Table~\ref{recongainTable} summarizes the
reconstruction gains for 
processing different size images in DWT
domain.
We can see that the time saved by skipping
RGB image reconstructing for 
inference accounts for over 
80\% of the total decoding time. 
The decoding time is based on an INTEL $ 7^{th} $ GEN CORE i7-7700HQ CPU. This gain improves with growing image size 
to allow better inference speed compared to the RGB domain.
	
	\begin{table}[!htbp]
		\vspace*{-5mm}
		\caption{Reconstruction gain at different image sizes.}
		\label{recongainTable}
		\centering
		\begin{tabular}{llll}
			\toprule
			Image size & decoding time (ms)     & recon. gain (ms) & recon. gain (\%)\\
			\midrule
			32$\times$32&25&20&80.0\\
			64$\times$64&31&25&80.6\\
			224$\times$224&109&91&83.5\\
			\bottomrule
		\end{tabular}
		\vspace*{-2mm}
	\end{table}
	
Figure~\ref{figtesttrain} compares the classification accuracy and the speed of inference and training for uncompressed images (i.e., $r=0$) of CIFAR-10 dataset. Each point in
 Fig~\ref{figtesttrain} represents the test accuracy and the speed of a particular model. 
To generate the results for a method, we changed the number of ResNet blocks to obtain better accuracy at the cost of training and inference time. 
To calculate inference and training speed, 
we \textbf{did not include} the reconstruction gain as explained in Section~\ref{recongain}. Inclusion of this
extra gain would shift the DWT curves 
more favorably to the right. We
gained approximately 2\% improvement in classification accuracy by applying our proposed augmentation techniques 
to the CIFAR-10 dataset. 
	
\begin{figure}[!htbp]
		\centering
		\hspace*{-5mm}
		\includegraphics[width=145mm,keepaspectratio,]{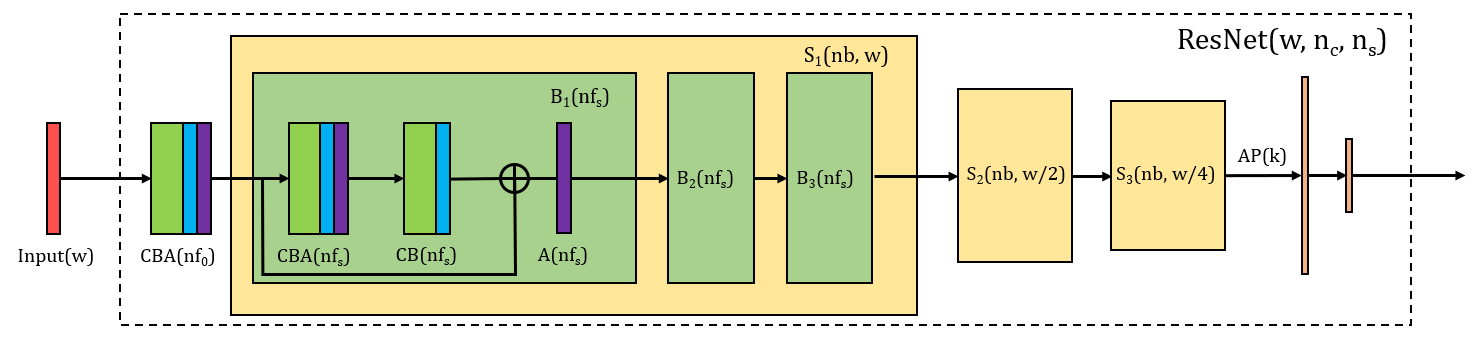}
\caption{ResNet(w, $n_{\rm{c}} $, $ n_{\rm{s}} $) represents a 
ResNet with $ n_{\rm{s}}=3 $ stacks, each stack $S(n_{\rm{b}},w)$ 
consists of $ n_{\rm{b}} $ identical blocks for input 
image size w and $n_{\rm{c}}$ channels. 
Each block $B(nf_{\rm{s}})$ consists of 2 convolution layers 
with $ nf_{\rm{s}} $ filters for the residual path and an identity connection followed by an activation function. 
After $S_3$, the output is average-pooled (AP) with 
kernel size $k$ and a dense layer is formed. 
$S_2$ and $S_3$ give additional convolution layer in the identity connection of the first block to match the down sampled input.}
		\label{figresnet}
		\vspace*{-5mm}
	\end{figure}

Table~\ref{resnettable} and Fig.~\ref{figresnet}
describe the ResNet model settings corresponding to 
the result points $a,b,c,d,e$ and $f$ in Fig.~\ref{figj2k} and Fig.~\ref{figtesttrain} 
for the CIFAR-10 dataset. 
Note that the numbers of convolution layers of the models 
used for DWT are lower than those used for
RGB. Although DWT models used more kernels requiring more 
parameters, the RGB input width is 2 times larger and
evens out the time spent by DWT models on more kernels.
We used an initial learning rate of 0.001 which is reduced 
progressively by $1/10$ at 80, 120 and 160 over 200 epochs.

	\begin{table}[!htbp]
		\vspace*{-3mm}
		\caption{Parameters for the ResNet models for RGB and CDF 9/7 inputs of CIFAR-10.}
		\label{resnettable}
		\centering
		\begin{tabular}{lllllll}
			\toprule
			\multicolumn{1}{c}{Parameter/Domain} & \multicolumn{3}{c}{RGB} & \multicolumn{3}{c}{CDF 9/7}              \\
			\cmidrule(r){1-1}\cmidrule(r){2-4} \cmidrule(r){5-7} 
			Model &a     & b     & c &d&e&f \\
			\midrule
			w&32&32&32&16&16&16\\
			$ n_{\rm{c}} $&3&3&3&12&12&12\\
			$n_{\rm{b}}$&4&3&2&3&2&1\\
			$nf_0$&16&16&16&64&64&64\\
			$nf_1$&16&16&16&64&64&64\\
			$nf_2$&32&32&32&96&96&96\\
			$nf_3$&64&64&64&144&144&144\\
			$k$&8&8&8&4&4&4\\
			no of CONV layers&27&21&15&21&15&9\\
			no of parameters (M)&0.37&0.27&0.18&1.79&1.17&0.55\\
			\bottomrule
		\end{tabular}
		\vspace*{-5mm}
	\end{table}
	
For ImageNet experiments, we used a ResNet with 32 convolution layers with bottleneck architecture 
described in \cite{he2016deep} for both RGB and DWT domains. This resulted in 3\% accuracy gain over regular ResNet architecture for the DWT domain 
and 0.3\% improvement for the
RGB domain. The initial learning rate is 0.001 which is reduced by $1/10$ at 30, 60 and 90 over 95 epochs,
respectively. 
	\begin{table}[!htbp]
		\vspace*{0mm}
		\caption{Results of CIFAR-10 based on models `a' and `d'.}
		\label{cifar10Table}
		\centering
		\begin{tabular}{lllll}
			\toprule
			parameter & RGB     & CDF 9/7Aug (ours)&CDF 9/7\\		\midrule
			Test Acc. ($\%$)&91.70$(\pm$0.07 )&\textbf{91.92}($\pm$0.11)&89.41($\pm$0.07)\\
			No of CONV layers&27&21&21\\
			Training rate/epoch (images/s)&862&\textbf{1000}&1000\\
			Inference rate (images/s)&4108&\textbf{4283}&4259\\
			\bottomrule
		\end{tabular}
		\vspace*{-4mm}
	\end{table}
	
We summarize the best classification models 
obtained for CIFAR-10 and Tiny ImageNet datasets in Table~\ref{cifar10Table} and Table~\ref{imageNetTable},
respectively.
We repeated each experiment 3 times to compute
the average and standard deviation of accuracy rates. 
Both data sets validate our claim of
faster training and inference for 
JPEG2000 compressed images even without considering the reconstruction gain which is considerably large and grows with image size. 
Similarly, both datasets demonstrate
the effectiveness of our proposed augmentation 
techniques for DWT domain images. 
CIFAR-10 shows over 2.5\% 
and Tiny ImageNet achieves
over 1\% and 1.5\% of top-5 and top-1 accuracy
improvement, respectively. 
	
	\begin{table}[!htbp]
		\vspace*{-4mm}
		\caption{Results of Tiny ImageNet.}
		\label{imageNetTable}
		\centering
		\begin{tabular}{lllll}
			\toprule
			parameter & RGB     & CDF 9/7Aug (ours)&CDF 9/7\\		\midrule
			Top 5 test Acc. ($\%$)&89.06($\pm0.03$ )&\textbf{89.08}($\pm0.02$)&87.92($\pm$0.07)\\
			Top 1 test Acc. ($\%$)&67.35($\pm$0.11 )&\textbf{67.56}($\pm$0.09)&65.78($\pm$0.36)\\
			No of CONV layers&40&31&31\\
			Training rate/epoch (images/s)&670&\textbf{694}&694\\
			Inference rate (images/s)&1865&\textbf{1881}&1881\\
			\bottomrule
		\end{tabular}
		\vspace*{-2mm}
	\end{table}

Consider the accuracy of the proposed work in RGB domain.
Both CIFAR-10 and Tiny Imagenet results show top-1
accuracy improvement of 0.2\% on average. 
Overall, both experiments suggest that RGB image reconstruction from j2k stream is not necessary for CNN 
classification. Including the reconstruction gain, 
DWT domain models can perform more than $\times$2 times faster over RBG domain. Although faster decoders 
implemented on GPUs may reduce
the gain, but as shown in Table~\ref{recongainTable}, 
this gain grows with image size.

\subsection{Experiments over Bandwidth Constrained Channels}

\begin{figure}[!htbp]
		\centering
		\vspace*{-7mm}
		\hspace*{-5mm}
		\subfloat[]{{\includegraphics[width=8cm]{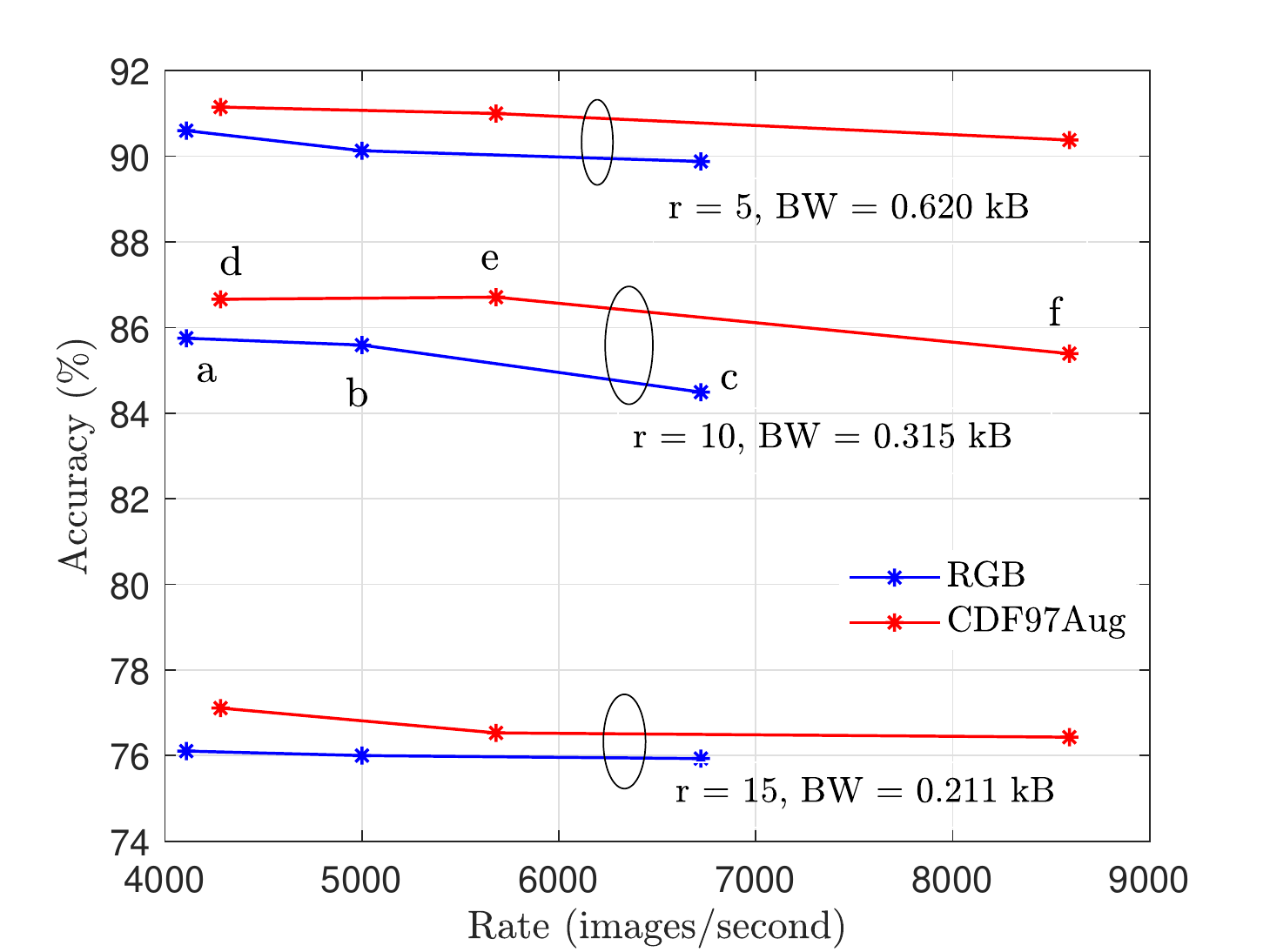}}}\hspace*{-6mm}
		\subfloat[]{{\includegraphics[width=8cm]{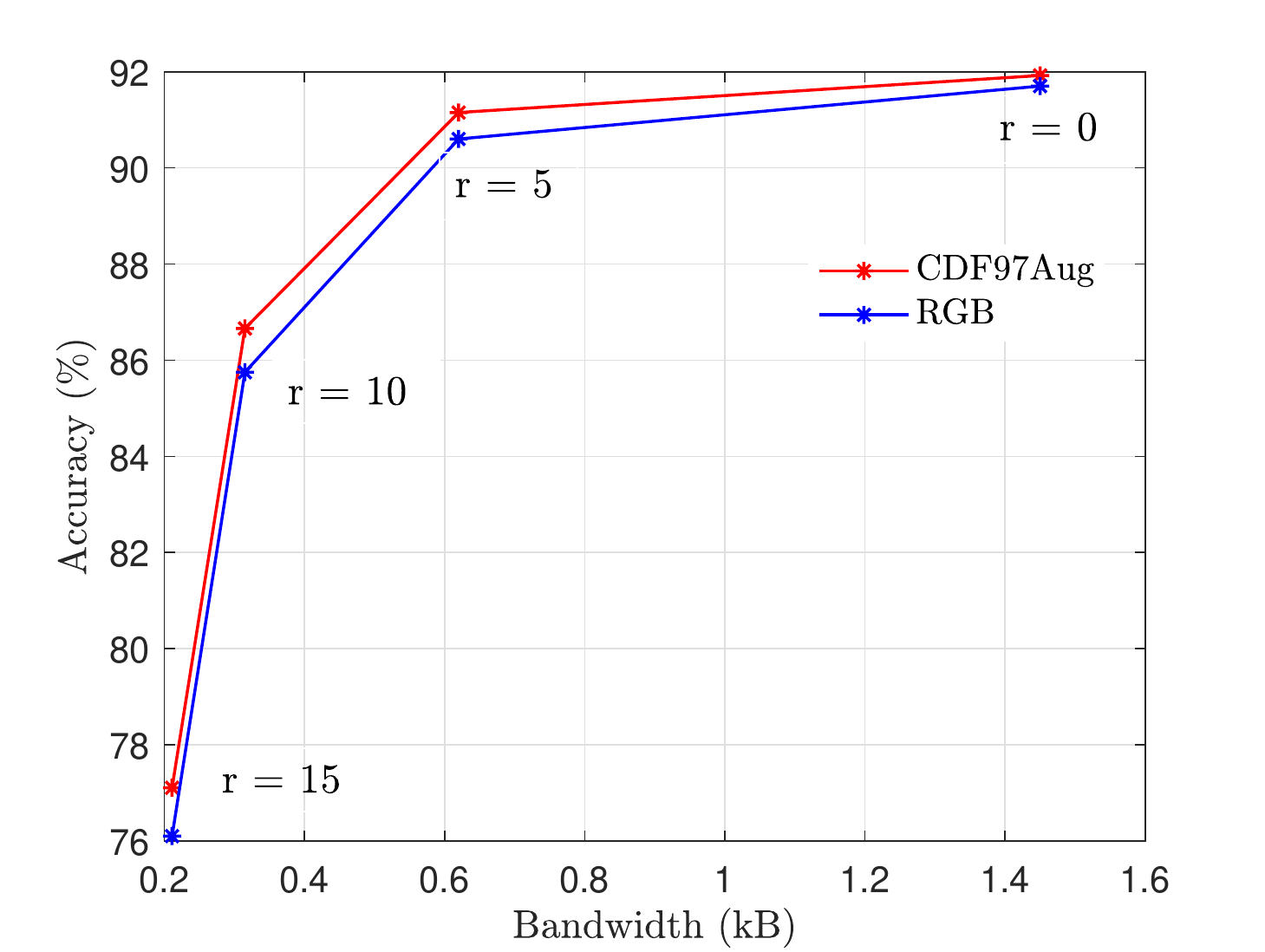}}}
		\vspace*{-2mm}
\caption{(a).
Models designed for CDF 9/7 DWT coefficients using the proposed augmentations are faster and 
more accurate under limited bandwidth. 
Parameter `r' in JPEG2000 codec adjusts compression ratio and 
`BW' is the bandwidth in terms of average image size. 
(b). Effect of channel bandwidth. RGB uses model `a' and CDF 9/7 uses model 'd'. This result is consistent with
result from DB1 wavelets in \cite{chamain2018QuanNet}}
		\vspace*{-3mm}
		\label{figbw}
	\end{figure}

Figure~\ref{figbw} shows how the classification accuracy behaves under bandlimited channels. Results in
Fig.~\ref{figbw}(a) confirm
faster and accurate classification for DWT domain 
models for channels with limited bandwidth. 
Results given by Fig.~\ref{figbw}(b)
show the change of accuracy for a particular model 
for different bandwidth constraints. These results
for JPEG2000 encoder that uses CDF 9/7 DWT coefficients
are consistent with the tests results from
DB1 wavelets given in \cite{chamain2018QuanNet}.
	
\begin{figure}[!htbp]
		\vspace*{-4mm}
		\centering
		\includegraphics[width=85mm,keepaspectratio,]{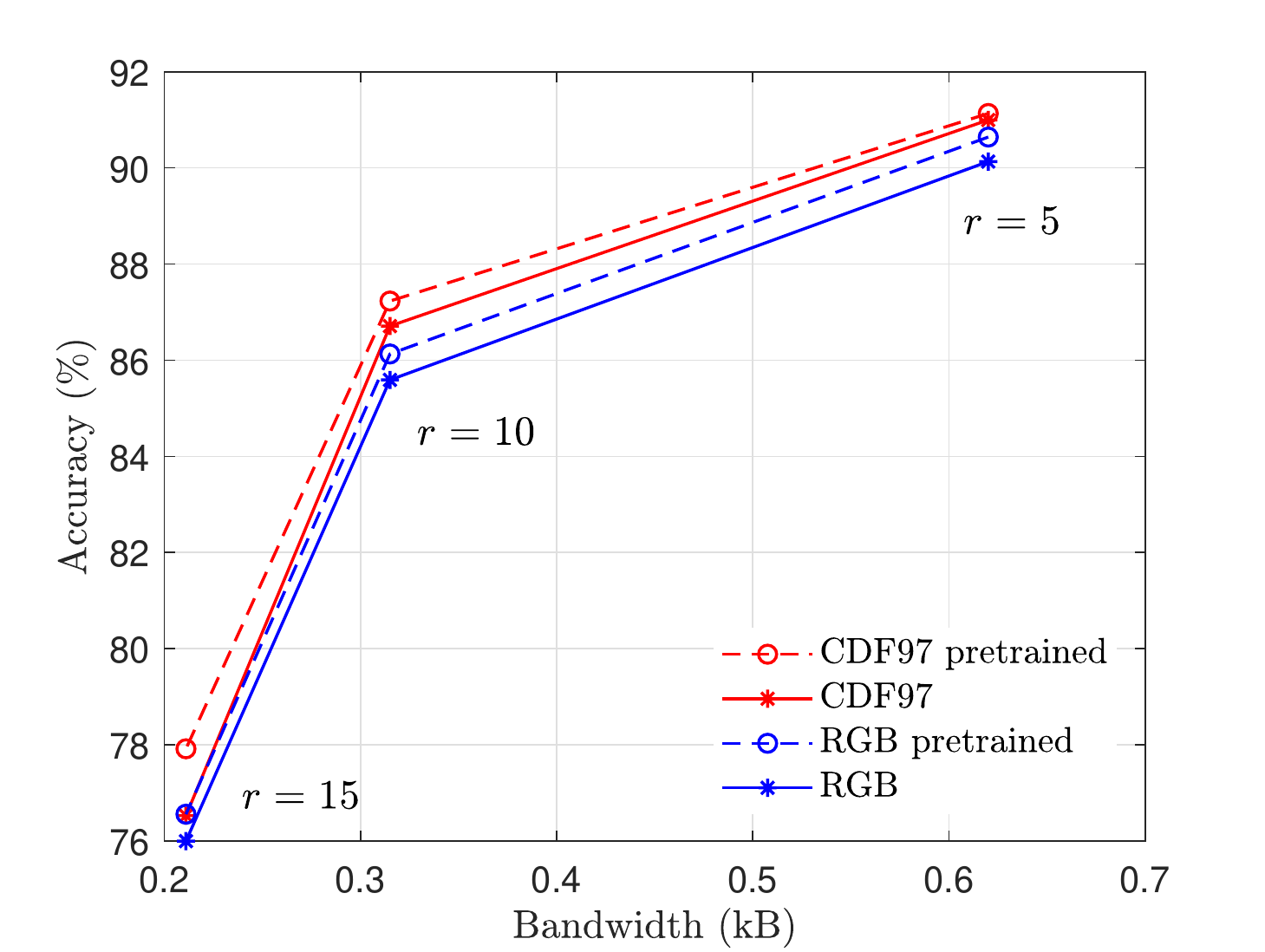}
		\vspace*{-2mm}
\caption{Fine tuning for $r=5,10,15$ using the 
pretrained DWT model 
for $r=0$ 
reduces training time by 75\% and improves accuracy. This observation is consistent with pre-training in RGB domain. CDF and RGB inputs use model `b' and `e' respectively.}
		\label{fig5}
		\vspace*{-3mm}
	\end{figure}

Fine tuning a base model for limited bandwidth can reduce
training time. Figure~\ref{fig5} shows that the 
pre-training is feasible and achieves a modest
accuracy gain for both RGB and CDF 9/7 DWT. 

	\vspace*{-3mm}

\section{Conclusions}
	\vspace*{-1mm}
This work investigates cloud-based deep CNN image
classification in congested communication 
networks. We
proposed to directly train deep CNN
classifier for JPEG2000 encoded images by
using its DWT coefficients in j2k streams.
We achieved better accuracy and simpler
computation by using shallower 
CNNs in the DWT domain.
We further introduced new augmentation transforms 
to develop CNN models that are robust to common
communication bandwidth constraints in cloud
based AI applications. 
\newpage
	\small
	\bibliographystyle{IEEEbib}
	\bibliography{Chamain_Draft_v4DZ}
\end{document}